\documentclass[fleqn,10pt]{wlscirep}
\usepackage[utf8]{inputenc}
\usepackage[T1]{fontenc}
\title{Controlling the Solo12 Quadruped Robot with Deep Reinforcement Learning}

\author[1,2,*]{Michel Aractingi}
\author[1]{Pierre-Alexandre L\'eziart}
\author[1]{Thomas Flayols}
\author[2]{Julien Perez}
\author[2]{Tomi Silander}
\author[1]{Philippe Sou\`eres}

\affil[1]{LAAS-CNRS, Universit\'e de Toulouse, Toulouse, 31400, France}
\affil[2]{NAVER LABS Europe, Meylan, 38240, France}

\affil[*]{michel.aractingi@naverlabs.com}

\begin{abstract}
Quadruped robots require robust and general locomotion skills to exploit their mobility potential in complex and challenging environments. In this work, we present an implementation of a robust end-to-end learning-based controller on the Solo12 quadruped. Our method is based on deep reinforcement learning of joint impedance references. The resulting control policies follow a commanded velocity reference while being efficient in its energy consumption and easy to deploy. We detail the learning procedure and method for transfer on the real robot. We show elaborate experiments. Finally, we present experimental results of the learned locomotion on various grounds indoors and outdoors. These results show that the Solo12 robot is a suitable open-source platform for research combining learning and control because of the easiness in transferring and deploying learned controllers.
\end{abstract}
\begin{document}

\flushbottom
\maketitle
%
%
\thispagestyle{empty}

\section*{Introduction}
\label{intro}

\noindent Legged robots can traverse challenging, uneven terrains. The interest in the design and control of legged robots has resurged due to the development of many quadruped platforms such as the  Mini-Cheetah\cite{Katz2019}, HyQ\cite{Semini2011}, ANYmal\cite{Hutter2016}, Solo\cite{Grimminger2019}, Spot Mini\cite{bostondyn} and Laikago\cite{unitree}. These platforms serve as suitable test-benches for control and locomotion research. Finding the right way to control such systems is crucial to fully exploit quadruped mobility. In this paper, we conduct our experiments using the Solo12\cite{Leziart2021} robot which is a recent alternative platform that provides a reliable low-cost open-access quadruped within the Open Dynamic Robot Initiative\footnote{https://open-dynamic-robot-initiative.github.io/}. 

 \begin{figure}
     \centering
     \includegraphics[width=\linewidth]{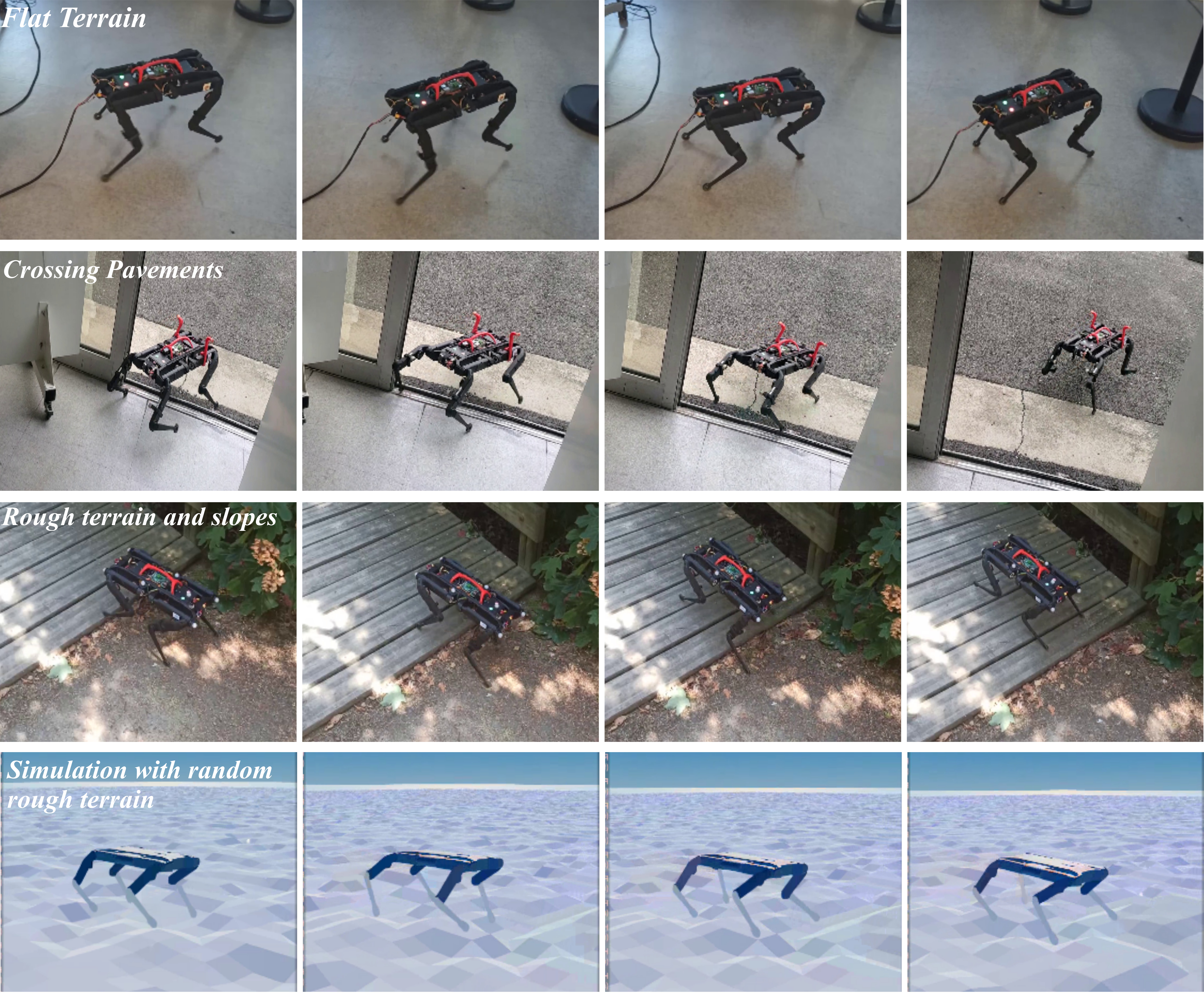}
     \caption{Snapshots of the Solo12 quadruped in  real settings and in simulation driven by a reactive controller learned through deep reinforcement learning. With learned controllers, the robot can traverse various outdoor environments with slopes and rough ground, full video: https://youtu.be/t-67qBxNyZI}
     \label{fig:snapshots}
 \end{figure}
 
Many control methods based on motion planning and trajectory optimization have been proposed for quadrupeds. Winkler et al.\cite{Winkler2015} suggest using a tree search to plan the body path and footsteps positions in the environment for the HyQ robot\cite{Semini2011}. Bellicoso et al.\cite{Bellicoso2018} show a ZMP-based motion planner for executing dynamic transitions between gaits on the ANYmal robot\cite{Hutter2016}. The approaches proposed by DiCarlo et al.\cite{DiCarlo2018} and Kime et al.\cite{Kim2019} use model-predictive control (MPC) on a centroidal model to plan the base trajectory and ground reaction forces of the feet in contact for the Mini-cheetah\cite{Katz2019}. Kim et al.\cite{Kim2019} also propose a whole body control formulation that outputs the necessary low-level control in order to track the base trajectory on a shorter-time horizon. L\'eziart et al. \cite{Leziart2021} implement a similar MPC-based approach for Solo12\cite{Grimminger2019} while simplifying solutions for the computation of the whole-body control. While all these methods produce robust dynamic controllers,  they often require some aspects of control, such as gait, feet trajectories, body height and orientation etc., to be determined by hand-tuned parameters that are hard to adapt for all the different environments a quadruped might encounter in the real-world. These controllers often rely on models that are hard to design and observe in many situations. Furthermore, these methods are computationally heavy at run time and often require laborious effort to set up. 
 
In contrast to optimization methods, data-driven methods that are based on learning can be used for designing controllers. 
Specifically, reinforcement learning (RL) is an alternative approach for obtaining highly performant agents that act in their environment in which the dynamics and transitions are modeled as a Markov decision process~(MDP)\cite{Suttonbook}. 
There are many early examples of applying RL to robotic tasks such as manipulation\cite{Gullapalli1995,Peters2008, Kober2010, Kalakrishnan2012} and locomotion\cite{Benbrahim1997, Kohl2004}. 
However, RL used to be hard to scale and was often limited to solving small sub-problems in the control pipeline in which most of the components were hand-designed. 
With increased computing power and recent evolution of deep learning methods that use large scale neural networks, we can now solve problems requiring high-dimensional data\cite{Lecun1998, Krizhevsky2012, Lecun2015}. Deep RL combines neural networks with RL algorithms to learn value function approximations\cite{Mnih2013, Mnih2015, Koutnik2013} and/or, directly, policies\cite{Levine2013, Schulman2015, Schulman2017}. Using images from camera, deep RL has been successfully applied for manipulation tasks such as object insertion, peg in a hole\cite{Levine2015}, and reaching and grasping objects\cite{Kalashnikov2018}.

In recent works, deep RL has been applied to quadrupeds\cite{Hwangbo2019} and bipeds\cite{Li2021} for the purpose of learning end-to-end controllers. Hwangbo et al.\cite{Hwangbo2019} outline a general RL method for learning joint angle controllers from the base and joint states of the robot. The authors propose learning a model of the actuation dynamics of ANYmal\cite{Hutter2016} from real-data that can then be deployed in simulation, thus enabling the learned policies to transfer to the real-world. In the work by Miki et al.\cite{Miki2022}, the authors deploy a similar learning scheme and augment the action space with a central pattern generator (CPG) layer that produces a baseline walking gait pattern for the feet\cite{Hoyt1981}. Using proprioception and a LIDAR based reconstruction of the environment, the policy then learns to manipulate the CPG phase and joint angles to modify the gait. Similarly, Lee et al. \cite{Lee2020} learn a policy that modifies the phase and shift of CPG functions that determine the foot trajectories which are fed to model-based controller to produce joint angle control. Ji et al.\cite{Ji2022} propose learning a control policy through RL and a state estimation network with supervised learning that tries to predict state variables that are not measured on the real robot but are available in simulation and provide vital information for learning robust policies, e.g., feet contact states and linear velocity of the base. These works mostly rely on domain randomization techniques that add noise to the sensory input of the policy and to the dynamics of the simulation in order to learn policies that transfer to the real system. Recent work also proposed to learn different skills for the Solo8 robot through imitation learning of  sequences generated by trajectory optimization method~\cite{li2023cassi, Fuchioka2022-opt-mimic}. In our work, we focus on using RL to learn robust end-to-end controllers from scratch for the Solo12 robot. 



In this paper, we present an RL approach for learning robust controllers on the Solo12 robot\cite{Leziart2021}. We use similar RL techniques for learning locomotion while introducing curriculum processes at different levels and randomization schemes for zero-shot transfer to the real robot.  We detail our procedure for setting up the MDP components, i.e., state space, action space and reward function, along with the additional techniques required to make the learning converge and transfer to the real robot. We use proximal policy optimization (PPO)\cite{Schulman2017} as the RL algorithm. Finally, we present videos and plots to describe experimental results showing the execution of the learned locomotion by the Solo12 quadruped indoors and outdoors. Figure~\ref{fig:snapshots} depicts examples of Solo12 controlled by learned policies in simulation and real-world using two different joint angle configurations
Our main contributions are:
\begin{itemize}
    
    \item Detailed description and analysis of a deep RL method for learning controllers for the Solo12 that transfer to the real-robot. 
    
    \item Introduction and study of a realistic energy loss penalty for policy learning based on actuator friction and Joules losses identification. 
    
    \item Open-source implementation to make the work reproducible that is in line with the open-source mission of Solo12. \footnote{Code available at: https://github.com/Gepetto/soloRL}
    
    \item Intensive tests of the learned locomotion on the Solo12 quadruped indoors and outdoors. 

\end{itemize}

In the next section, we present notations and preliminaries for RL and MDPs.  After that, we explain our learning methods and notably the core components of the MDP, i.e., the state, actions, reward function and transfer methods. The Experiments-section showcases our results in simulation and with the real robot. Finally, we offer concluding remarks.

\begin{figure*}[t!]
    \centering
    \includegraphics[width=0.9\textwidth]{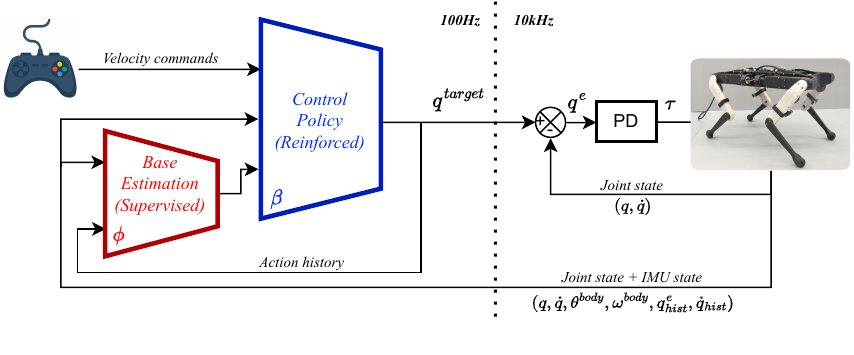}
    \caption{Summary of the control scheme shows the policy network, base estimation network and the low level control setup on the real robot 
    }
    \label{fig:neural-architecture}
\end{figure*}

\section*{Reinforcement Learning Preliminaries}
\label{sec:prelim}

\noindent We model the reinforcement learning (RL) environment as a Markov decision process (MDP) with continuous state and action spaces\cite{Suttonbook}. 
An MDP is defined by the tuple $(\mathcal{S}, \mathcal{A}, \mathcal{R}, \mathcal{T}, P_0)$, where $\mathcal{S} \subset \mathbb{R}^{d_S}$ is a set of states, and $\mathcal{A} \subset \mathbb{R}^{d_A}$ is a set of actions. 
In RL setting, only spaces $\mathcal{S}$ and $\mathcal{A}$ of the MDP are known to the learning agent. The agent starts by observing the initial state $s_0\in\mathcal{S}$ and it performs actions 
$a_t\in\mathcal{A}$ in the environment at discrete times indexed by $t\in\mathbb{N}$, after which it receives a stochastic reward $r_{t+1}\in\mathbb{R}$ and observes a new stochastic state $s_{t+1}$. 

The environment dynamics are described by a  transition  probability  distribution $\mathcal{T}: \mathcal{S} \times \mathcal{A} \times \mathcal{S} \rightarrow \mathbb{R}_+$,  such that $\mathcal{T}(s,a,s^\prime) = p(s^\prime \mid s,a)$ is the probability (density) that the next state is $s^\prime$ given that the current state is $s$ and that the action taken is $a$. $P_0$ is  the  initial  state  probability  distribution. Similarly, the stochastic reward $r\in\mathbb{R}$ after taking an action $a$ in a state $s$ and observing a state $s'$ next is governed by the function $\mathcal{R}:\mathcal{S} \times \mathcal{A} \times \mathcal{S} \times \mathbb{R} \rightarrow \mathbb{R}_+$ that defines the probability densities $p(r \mid s,a, s^\prime )$. While in general $\mathcal{R}$ is defined as a density, in our simulations the reward function is a deterministic function of $a$ and $s'$.



 To formalize the goal of learning, we define a  stochastic policy $\pi_{ \beta}(s, h, a) = p_{ \beta}(a_t = a \mid s_t = s, h_t = h_t^{k})$, parameterized by $ \beta$, that gives the probability density of taking an action $a$ given a state $s$ and a history $h^k$ of length $k$ where h contains parts of the past states and actions from $t-1$ to $t-k$ timesteps. In general it could contain the full states and actions of the last $k$ timesteps $h_{t}^{k}=(s_{t-k}, a_{t-k}, ... \: s_{t-1}, a_{t-1})$. However, in our work we found that we only need the history of the actions and joint states as will be disussed in the following section. 
The learning objective is to find the parameters $ \beta$ of the policy for which the expected  discounted sum of rewards 
$J( \beta) := \mathbb{E}[\sum_{t=1}^{H} \gamma^{t-1} r_t]$ is maximized.
In this expression $H$ is the horizon of the episode and $\gamma \in [0,1]$ is a discount factor. The expectation is taken over the stochastic policy, the initial state distribution, and the stochasticity of rewards and state dynamics. 


\section*{Method}
\label{sec:method}
\noindent Our goal is to define an RL method that can learn to control a Solo12 robot to follow a user-defined velocity command. The Solo12 quadruped is a 12 degrees of freedom version of Solo8\cite{Grimminger2019} that can be torque controlled. We will describe the design of our state space, action space and reward function in the following sections.

In general, our control policy is implemented as a neural network that takes the state as an input, and outputs the actions. The actions that define joint angle targets are then fed to a Proportional Derivative (PD) feedback controller in order to get the desired torques for commanding the robot joints. Figure~\ref{fig:neural-architecture} shows a summary of the control scheme in terms of the inputs/outputs of the control network and how it is deployed on the real robot. The estimation network in Figure~\ref{fig:neural-architecture} is trained with supervised learning to predict the linear velocity of the base. 
The control policy parameters are optimized using the proximal policy gradients objective (PPO)\cite{Schulman2017}.

\subsection*{State space}
\noindent The state space of the MDP is constructed from the proprioception of the robot, i.e., the sensory readings from the  joint encoders, and the inertial measurement unit (IMU). The state at time $t$ includes the base state and the joint state. The base state consists of the orientation $\theta^{\text{body}}_t \in \mathbb{R}^3$, linear velocity  $v^{\text{body}}_t \in \mathbb{R}^3$ and angular velocity $\omega^{\text{body}}_t \in \mathbb{R}^3$ of the body. The joint state consists of the joint angles $q_t \in \mathbb{R}^{12}$, joint velocities  $\dot{q}_t \in \mathbb{R}^{12}$ along with history of the joint target errors $ q^{e}_{hist, t} =  \{q^e_{t - j}\in \mathbb{R}^{12}\}^{j=1...N}$ (explained below) and joint velocities $ \dot{q}_{hist, t} = \{\dot{q}_{t - j}\in \mathbb{R}^{12}\}^{j=1...N}$. In our work $N=3$, i.e., the velocities and joint target errors from last three policy steps are stored and added to the state. We also include to the state $s_t$ the last two actions $\{a_{t - j}\in \mathbb{R}^{12}\}^{j=1...(N-1)}$. Finally, the 3D velocity command is also given as an input to the policy neural network. 

The orientation and angular velocity of the base can be provided by an IMU on-board the robot, which internally uses an Extended Kalman Filter (EKF) to estimate angular orientation from raw gyroscope and accelerometer sensor data. At each joint an optical encoder measures the joint angles, from which one can then compute the joint velocities. The joint target errors are the differences between the target joint angles conveyed to the PD controller and the measured joint angles, i.e., $q^e_t = q^{target}_{t-1} - q_{t}$. The error $q^e_t$ is related to a torque, and it implicitly provides rich information, such as the contact state of the feet with the ground, about the environment. The target errors also vary by terrain as the vertical foot position shifts if the terrain is not flat, which changes the resulting joint angles. Therefore, it is also crucial to add the last two actions of the policy to the state so that the learning can observe the change of the joint target errors for the similar actions which indicates a change in the terrain. 

The on-board IMU does not directly measure linear velocity, and estimating the velocity from accelerations often diverges over time due to sensor bias. Like Ji et al. \cite{Ji2022}, we propose training a separate state estimation network for estimating the base linear velocity from the IMU and joint encoder measurements. The state estimation network is trained through supervised learning and it receives as input the base orientation and angular velocity along with the joint angles, joint velocity, history of the past joint angle errors, joint velocities and actions. The output is a three-dimensional vector that estimates the linear velocity in the $x,y,z$ directions. Implementation details can be found in Experimental Results section. 

\subsection*{Action space}
\noindent The design of the action space can make a difference on the learning speed and policy quality. Peng et al.\cite{Peng2016} showed that direct torque control is harder to learn than joint position control in RL-based systems. Similar observations were made in the literature on learning  quadruped robots' locomotion\cite{Hwangbo2019, Kumar2021}. We also argue that torque control policies are harder to transfer than joint angle control policies, due to the fact that joint angle control is inherently stable after choosing appropriate impedance gains $K_p$ and $K_d$. While direct torque control can result in diverging motion, especially during the flying phases of the legs where the apparent joint inertia is low, the position-based impedance control forces the joints to behave like a spring damper system.

In this work, we propose learning a policy $\pi$ that outputs displacements of the reference joint angles with respect to the nominal pose of the robot, i.e., 
$\pi_{ \beta}(s_t) = \Delta q^{ \beta}_t$, where $\pi$ is implemented by the policy neural network parameterized by $ \beta$, and $s_t$ is the state input to the policy at time $t$. The target joint angles can then be computed as: 
\[
q^{target}_t = q_{init} + \lambda_q \Delta q^{ \beta}_t,
\]
where $q_{init}$ are robot's nominal joint  configuration around which the policy actions are centered. We define $\lambda_q$  as a constant that scales the output of the network before adding to $q_{init}$. Given $q^{target}_t$, we use a PD controller to compute the torques:
\[
\tau_t = K_p (q^{target}_t - q_t ) - K_d \dot{q}_t
\]
with the proportional and derivative gains $K_p$ and $K_d$.
It is important to note that using such a joint controller doesn't imply having a rigid position control. The reference angle $q^{target}_t$ should not be interpreted as positions to be reached, but rather as intermediate control variables. The resulting system is analog to elastic strings that pull the joint angles toward $q^{target}_t$.

\subsection*{Reward function}
\noindent The reward function defines the task. 
The main task in our work is to follow a given reference velocity. In order to get natural locomotion that can be deployed on the robot, one needs some constraints on the robot's pose, joint torques, joint velocities, etc. After each action $a_t$, the robot receives a reward $r_{t+1}$. We split our reward $r$ into one main positive term that rewards the tracking of the commanded velocity and several weighted penalty terms that act as negative costs in the reward. The values of the weights are listed in Table~3.
The reward terms and state variables below are implicitly indexed by the time step index $t$ but we only include this index when necessary for clarity. 

\subsubsection*{Command velocity tracking.} The reward $r_{vel}$ for following the command velocity is based on the squared Euclidean distance between the 3D vector $V_{x,y,w_{z}}$ consisting of the forward, lateral and yaw velocities of the body and the 3D velocity command $V^{cmd}$, i.e., 
\[
r_{vel}=c_{vel} e^{-\vert\vert V^{cmd}-V_{x,y,w_{z}}\vert\vert ^2}
\]
with coefficient $c_{vel}$ that scales the reward. 

\subsubsection*{Foot clearance penalty.} 
To encourage the robot to lift its feet high even when training on a flat surface, we use the foot clearance objective proposed by Ji et al.\cite{Ji2022}. 
Denoting the height of the i-th foot by $p_{z,i}$, we set a constant foot height target $p^{max}_{z}$ and define the foot clearance penalty as
\[
r_{clear} = c_{clear}\sum_{i=1}^{4} ( p_{z,i}-p^{max}_{z}) ^2\vert\vert \dot{p}_{xy,i}\vert\vert ^{0.5}, 
\]
where 
$\dot{p}_{xy,i}$ stands for the velocity of the foot $i$ in the $x,y$ direction so that the target is not active during the ground contact and it is approximately maximal in the middle of the swing phase. Scalar $c_{clear}$ is a weight for this penalty.

\subsubsection*{Foot slip penalty.} When a foot comes in contact with the ground, its $x,y$ velocity should be zero in order to avoid slipping. 
We define a foot slip penalty 
as 
\[
r_{slip} = c_{slip}\sum_{i=1}^{4} C_{i}\vert\vert \dot{p}_{xy,i}\vert\vert ^2,
\]
where $C_{i}$ is a binary indicator of the ground contact of the i-th foot, and  $c_{slip}$ is penalty weight.

\subsubsection*{Base orientation and velocity penalties.} The base pitch, roll and velocity in the $z$ direction should all be near zero to produce stable motion. With scalars $c_{orn}$ and  $c_{vz}$, we define this penalty as
\[
 r_{base} =c_{orn}(roll^2 + pitch^2)
 + c_{vz}V_z^2.
\]

\subsubsection*{Joint pose penalty.} We add a penalty on the joint angles in order to learn to avoid large joint displacement. We define this penalty as the deviation from the nominal joint angles at the initial state, as
\begin{align*}
r_{joint} = c_{q}\vert\vert q_t-q_{init}\vert\vert ^2
\end{align*}
with weight $c_{q}$. 

\subsubsection*{Power loss penalty.} For safety reasons and for saving energy, we would usually prefer to minimize the overall power consumption of the robot. The power loss term encapsulates the relationship between the torque and velocity at the joint level - we use the model proposed and identified by Fadini et al.~\cite{Fadini2021} which includes the heating by Joules loss in the motors $P_J$ as well as the losses by friction $P_f$.

We denote with $\tau_f$ the torque necessary to overcome the joint friction :
\begin{align*}
\tau_f = \tau_u\text{sign}(\dot{q}) + b\dot{q},
\end{align*}
where $q$, $\dot{q}$ are respectively the joint position and velocity.
The identified model parameters are the Coulomb friction $\tau_u = 0.0477$[Nm] and the viscous friction coefficient $b = 0.000135$[Nm$\cdot$s].

The two sources of power losses can then be expressed as
\begin{equation*}
P_f = \tau_f\dot{q}  \quad [W], \; \mathrm{ and }\; P_J = K^{-1}(\tau + \tau_f)^2  \quad [W], 
\end{equation*}
where $\tau$ is the joint output torque and $K = 4.81$[Nm$\cdot$s] is linked to the motor coil resistance and motor constant. 

The total power over joints and the penalty term used in the reward is taken as the sum over all joints:
\[
r_E = c_{E}\sum_{j=1}^{12} P_{f,j} + P_{J,j}
\]
with the weight $c_{E}$.

\subsubsection*{Action smoothness penalties.} To generate joint trajectories without vibrations and jitter,  we define a penalty on the first and second order differences in the joint angle values: 
\begin{equation*}
    r_{smooth} = c_{a1}\vert\vert q^{target}_t-q^{target}_{t-1}\vert\vert ^2 + c_{a2}\vert\vert q^{target}_t-2q^{target}_{t-1}+q^{target}_{t-2}\vert\vert ^2
\end{equation*}
with weights $c_{a1}$ and  $c_{a2}$.

\subsubsection*{Total reward.} The final reward is a weighted sum of the positive velocity tracking reward minus a sum $r_{pen}$ of all the penalties explained above:
\[
r_{total} = r_{vel} - r_{pen}. 
\]

\subsection*{Domain and dynamic randomization}
\noindent In order to learn policies that transfer to the real robot, we have to identify and bridge the sim-to-real gap. We decided to use domain randomization techniques by adding noise to the state and randomizing some aspects of the simulator dynamics. Table~1 
shows the noise models used for each element in the state and dynamics. For the dynamics, we found that for Solo12 it was enough to randomize the gains of the PD controller in order to learn policies that adapt to some stochasticity in the low level control that can come from many factors. This is in contrast to previous work on ANYmal and Mini-cheetah where more randomization is needed for the center of mass, mass of the body and links, positions of the joints and motor friction\cite{Ji2022, Hwangbo2019, Lee2020, Miki2022}.
Randomizing the state is essential in order to overcome sensory noise. Our results show that one can learn a transferable policy on Solo12 using this simple randomization strategy.

{\setlength{\tabcolsep}{1.0em}
\begin{table}[ht!]
\centering

\begin{tabular}{| c | c |} 
 \hline
 \multicolumn{2}{| c |}{Observation Noise}\\
 \hline\hline
 $\theta_{\text{body}}$ & $U^{3}(-0.05, 0.05)$ \\
 \hline 
 $\omega_{\text{body}}$ & $U^{3}(-0.10, 0.10)$ \\ 
 \hline
 $v_{\text{body}}$ & $U^{3}(-0.10, 0.10)$ \\ 
 \hline
 $q$ & $U^{12}(-0.05, 0.05)$ \\ 
 \hline
 $\dot{q}$ & $U^{12}(-0.50, 0.50)$ \\ 
 \hline\hline
  \multicolumn{2}{| c |}{Dynamics Noise}\\
 \hline\hline
 $K_p$ & $U(-1.0, 3.0)$\\
 \hline
 $K_d$ & $U(-0.1, 0.1)$\\
 \hline
\end{tabular}
\label{table:rand}
\vspace{2mm}
\caption{Uniform noise for each of the state observations \\ and PD controller gains.}
\end{table}
}

\subsection*{Curriculum learning}
\noindent \subsubsection*{Reward curriculum.} Due to the elaborate penalty terms of the reward function, we observe that the agent may learn to neglect the positive reinforcement signal from following the command velocity and learn to stand still, since this  optimizes several penalty terms in the reward. In order to bypass this problem, we introduce a linear curriculum on the reward. Curriculum learning is a popular method that introduces easier tasks to learn at the start of training and gradually increases the level of difficulty as training progresses\cite{Bengio2009}. Like Hwangbo et al.\cite{Hwangbo2019}, we multiply the cost terms of the reward function by a curriculum factor $k_c\in [0, 1]$ that is equal to zero at the start of the training and slowly increases up to one through the training iterations. The reward function becomes $r_{total} = r_{vel} -  k_c r_{pen}$.
This way we first train the agent to follow the command velocity in any manner before emphasizing the cost terms in the reward in order to refine locomotion.

\subsubsection*{Noise curriculum.} We also propose a curriculum on the injected noise for randomizing the state and dynamics. We found that decoupling the curriculum of the reward and randomization works better. Therefore, the sampled noise in Table~1 
is multiplied by another curriculum factor $k_{c,noise}\in [0.0, 1.0]$ that is increased at a slower pace than $k_c$.

\subsubsection*{Terrain curriculum.} We introduce rough terrains at the end of training to learn from more complex interactions when the ground is not flat. This helps in refining the robot's locomotion in terms of lifting all feet equally in order to keep balance. At the last 1000th training iteration, we start sampling random heightmaps at the start of the episodes.
We also lower some PPO parameters to perform more conservative updates to the policy in order to avoid catastrophic forgetting\cite{FRENCH1999128} of locomotion on flat terrain once the rough terrains are introduced and the training data distribution changes. The PPO parameter values before and after  introducing the rough terrains are listed in Table~2,
 we refer to Schulman et al.\cite{Schulman2017} for a description of these parameters.  
\begin{table}[t!]
\centering
\begin{tabular}{| c | c | c |} 
 \hline 
 PPO parameters & Flat terrain & Non-flat terrain\\
 \hline
 Clip ratio & 0.200 & 0.050\\
 \hline
 Gradient norm clip & 0.500 & 0.300\\
 \hline
 Entropy coefficient & 0.010 & 0.000\\
 \hline
 Learning rate & 0.005 & 0.001 \\
 \hline 
\end{tabular}
\label{table:ppo}
\vspace{2mm}
\caption{PPO parameters when training on\\ flat terrain and non-flat terrain.}
\end{table}

\section*{Experimental Results} 

\noindent In this section, we analyze the locomotion produced by our learned control policies. We test both symmetric ($\:\overline{><}\:$) and non-symmetric ($\:<\!\!\!\!\overline{~<}\:$) poses of the legs with the policy being able to learn both successfully. We display results about velocity tracking and energy consumption of the learned controller. Successful real robot transfer experiments are conducted and discussed in the following sections. 

\begin{table}[!ht]
\centering
\begin{tabular}{| c | c | c | c | c | c | c | c | c |} 
  \hline
 $c_{vel}$ & $c_{clear}$ & $c_{slip}$ & $c_{orn}$ & $ c_{vz}$ &
  $c_{q}$ & $c_{E}$ & $c_{a1}$ & $c_{a2}$ \\
  \hline
 $6.0$ & $20.0$ & $0.07$ & $3.0$ & $1.2$ & $0.5$ & $2.0$ & $2.5$ & $1.5$\\
  \hline
\end{tabular}

\label{table:weights}
\caption{Reward terms' weights.}
\vspace{-5mm}
\end{table}

\label{sec:experiments}
\subsection*{Implementation details}
\noindent The control policy is implemented as a multi-layer perceptron 
 with three hidden layers of sizes 256, 128 and 32 with Leaky~ReLU activations between each layer. 
The control policy runs at a frequency of 100Hz.
We use the Raisim simulator\cite{raisim} for training. The simulator frequency is set at 1kHz which means that the PD control between each RL step is executed ten times. On the real-robot we have a low-level loop at 10kHz for communicating with the actuators, but the policy network is still queried every 0.01 seconds (see Figure~\ref{fig:neural-architecture}). In simulation, 300 different versions of the robot are run in parallel processes in order to collect diverse data faster. The value of the PD control gains are $K_p=3$ and $K_d=0.2$ respectively. On the robot, the computation of actions from states only takes $10~\mu s$ on a Raspberry Pi 4 which makes this approach particularly appealing due to its simple setup and high computational speed. 

 The state estimation network is also a multi-layer perceptron with two hidden layers of sizes 256 and 128 with Leaky~ReLU activations and a three-dimensional output corresponding to the linear velocity. To train the state estimation network, we run the learned policy in simulation to collect a dataset of states, without linear velocity, that are the input to the state estimation network and the linear velocities that will be its output. We found that a dataset of 50,000 samples (policy steps) is enough to train the estimation network to a good accuracy. In Ji~et~al.~\cite{Ji2022}, the authors propose to learn both networks (estimation and control) simultaneously. In our experiments, we didn't observe any advantage when training both networks together and decided to train the estimation network after the control policy in order to not slow down the RL training due to the overhead from performing supervised learning every few RL iterations.
 The data is collected with the random noise added to the observations and PD gains along with randomizing the terrains between rough and flat. We train on a  supervised cost to minimize the mean squared error loss using the Adam optimization algorithm\cite{Kingma2014}. 
 
 \begin{figure*}[!ht]
     \centering
     \includegraphics[height=8cm,width=0.95\linewidth]{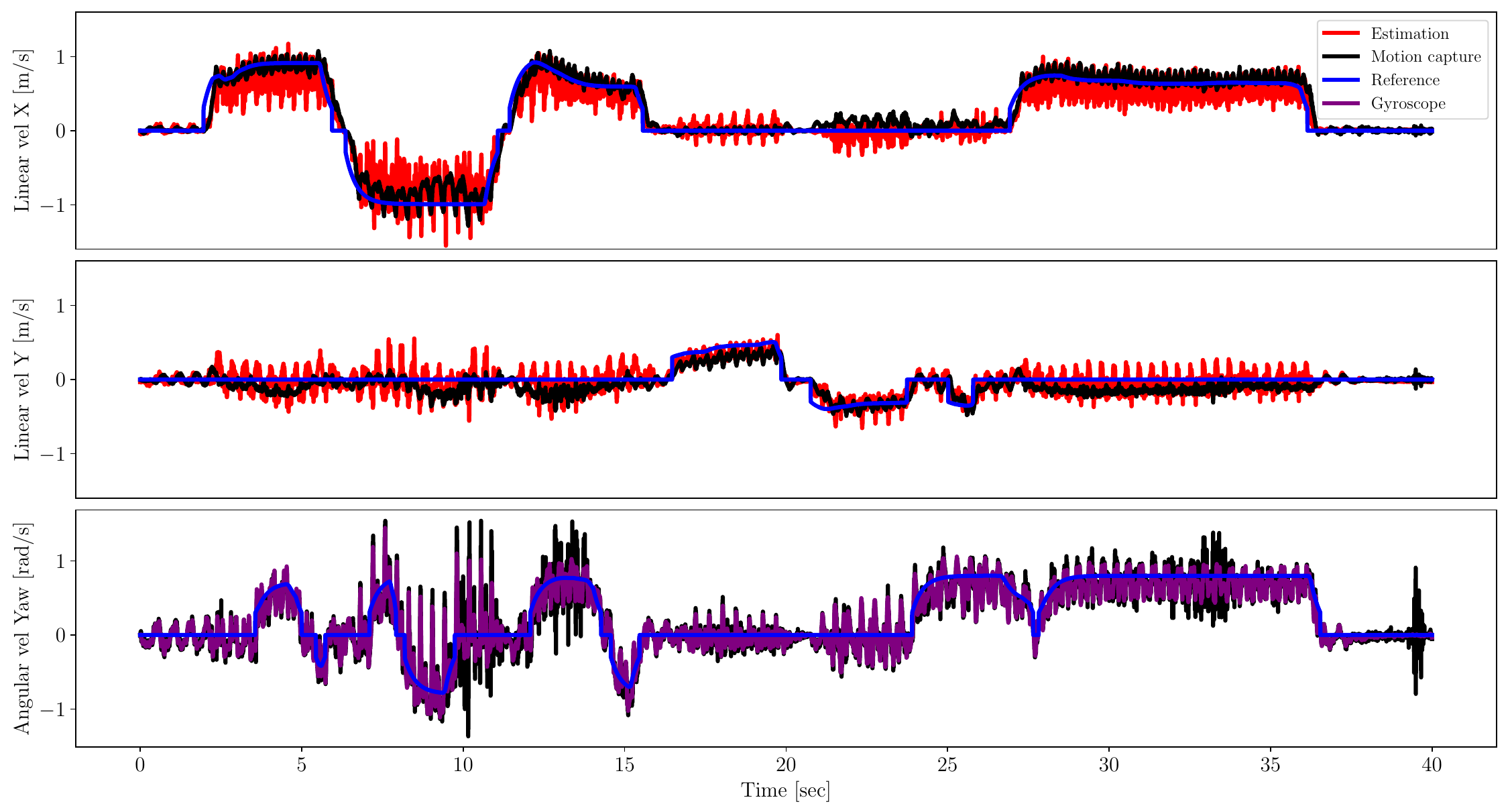}
     \caption{Plot of the 3D velocity command controlled by a gamepad to command the real robot in blue. The red curve plots the output of the state estimation. The black plot is the motion capture of the real solo12 to convey the ground-truth base velocity. The purple plot is the yaw velocity estimate from the gyroscope in the IMU. The x-axis shows time in seconds}
     \label{fig:vel_track}
 \end{figure*}

We use the objective from PPO\cite{Schulman2017} to train the policy network. This is done in an actor-critic setup where, in addition to the policy network (actor), we train another network (critic) that learns to map the state to a single scalar value that estimates the desirability of the state. This scalar value is commonly used for reducing the variance of the RL objective\cite{Schulman2017}. In each training episode, the policy is run for 100 steps (= 1 second of real-time) to collect data for optimizing the objective. The episode ends if the body of the robot comes in contact with the ground. Even though locomotion is not an episodic task with  a natural endpoint and the episode is not reset between each training epoch, we choose to introduce random resets at the beginning of some episodes since this appears to stabilize training. At the start of each episode, a random velocity command is sampled and then scaled by the noise curriculum factor so that  the network starts learning gradually from one low velocity towards higher ones. The initial state at the start of each episode is set at the nominal joint pose $q_{init}$ with zero joint velocity. We use the \texttt{stable-baselines}\cite{stable-baselines3} open source implementation of the PPO algorithm. 

As mentioned before, at the beginning of training the ground is flat, but in order to learn more robust policies, we gradually introduce some non-flat terrains by sampling random height values for points in a regular grid. At the last 1000th training iteration, 80\% of the parallel processes start sampling non-flat terrains. We found that we need around 10,000 training iterations which equates to 300 million collected samples with 300 parallel processes. 

Table~3
shows the coefficient values that are used to scale each term in the reward function. Along with choosing the right values of the weights, we choose the desired  maximum foot height in the foot clearance reward to be $p_{z}^{max}=6cm$. We scale the output of the policy network, with scalar $\lambda_q=0.3$ before integrating towards the target joint angles. 

\subsection*{Velocity tracking}
\noindent We first judge the quality of the learned controller by its ability to follow the reference velocity in the forward, lateral and yaw directions. During training, we randomly sample the velocity vector based on the following uniform distributions: $V_x \sim U(-1.5, 1.5)$, $V_y \sim U(-1, 1)$ and $W_z \sim U(-1, 1)$. As mentioned before, these values are scaled by $k_{noise}$ in order to start learning with low velocities before gradually increasing the range of sampled velocities. 

Figure~\ref{fig:vel_track} shows the velocity plots of a random walk recorded while guiding the robot with the gamepad across the room.  The blue lines plot the reference velocity command in three directions. The black lines represent the robot's body velocity estimation from motion capture data. The red lines in the first two plots are the state estimation network's velocity estimates in the $x$ and  $y$ directions. From the plots, we see that the real robot is able to follow the commanded velocity well, as indicated by the alignment between the motion capture plots --~which provides ground-truth values~-- and the reference command plots.   
The velocity predictions from the state estimation network are similar to the ones from motion capture, while being more noisy. The noise in the prediction, that is given as an input to the control network, does not appear to downgrade the performance of the controller. Indeed, this robustness to noisy estimation is expected as noise is added to the linear velocity input during training.

\begin{figure}[!t]
     \centering
     \includegraphics[width=0.95\linewidth]{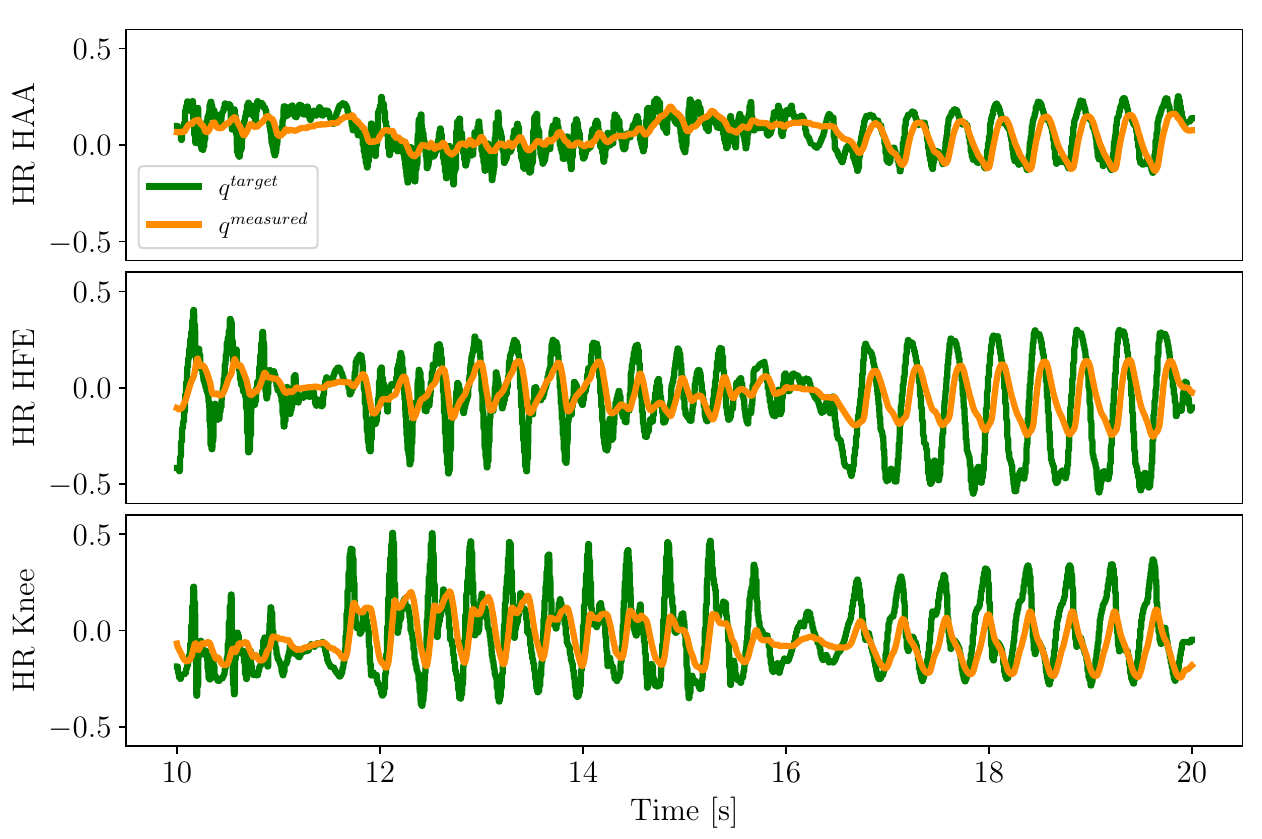}
     \caption{Plot of the desired joint angle command vs. the measured joint angles over a random run for the hind right leg. HFE stands for Hip-Flexion-Extension and HAA stands for Hip Aduction-Abduction}
     \label{fig:q_track}
 \end{figure}

Figure \ref{fig:q_track} shows the plot of the hind right joint angle target vs. the measured joint angles for the same random run. We observe that the target joint angles are not reached. 
The difference between the command and the achieved angles showcases the nature of the soft impedance control which, resembles elastic strings where the desired joint velocity is zero. Similar behaviour is observed for the other legs.

\subsection*{Energy consumption}
\noindent In order to verify the usefulness of the proposed power loss penalty in the reward function, we run several experiments while varying the power loss weight $c_{E}$ in the reward  and observe its effect on the learned policy. We run the policies in simulation for five seconds for the maximum forward velocity command of $1.5 [m/s]$. This test focuses on a rapid and dynamic task that would require most energy. 

Table~4
lists the effect of $c_E$ on the average power consumption, velocity error and base height during the test task. We first observe that the increase of $c_E$ decreases the power loss. This confirms that the power term on the learned policy makes intuitive sense and that it can be tuned to learn locomotion with different power profiles. We found for higher values for $c_E > 10$ the reward is ill-defined and training fails. 

Increasing the weight $c_E$ makes the policy prioritize optimizing the power loss rather than other rewards such as velocity tracking. We observe this effect in the table as the velocity error increases when using policies that have learned to consume less power due to higher $c_E$. The velocity error column contains the $l_1$ norm of the difference between the desired velocity and the achieved velocity. Note that even though the error increases, we see a big decrease in the consumed power, which would make the policies with $c_E\in [3, 4]$ an attractive option since the robot would have slightly less accurate velocity tracking but still save more than 30\% on the consumed power.

The base height could be another indicator of energy efficiency, since standing on straighter legs  requires less power. In Table~4
we list the body height as a function of $c_E$, and observe a gradual $2\,cm$ increase in the base height when $c_E$ increases from zero to ten. Beyond $c_E=10$ the RL ceases to produce good policies as mentioned before.   

\subsection*{Power vs. torque penalty}
\noindent
In previous work\cite{Hwangbo2019, Miki2022, Ji2022} penalty terms on the torque magnitude, joint velocity magnitude and joint accelerations are used in the reward.
We trained several policies using these penalty terms to compare with the proposed power cost. The last row in Table~4 
shows the power loss vs. velocity error for the policy trained with those penalties. The learned policy is less energy efficient than most of the policies that have the power term with high variance between the policies. In practice, we found it easier to tune a single power weight during experimentation rather than tuning three separate weights for torque, velocity and acceleration terms with different units. The power loss formula expresses the relationship between the torque and the velocity by effectively combining the other three penalties into a one single physical and coherent term.

\begin{table}[ht!]
\centering
\begin{tabular}{| c | c | c | c |} 
 \hline 
 $c_E$ & 
  Power [W] & Velocity error [m/s] (\%) &  Base height [m] \\
 \hline
  0.0 & $17.7$ 
  & $0.079 \pm 0.054$ (7.9\%)  & $0.23 \pm 0.004$ \\
 \hline
   0.1 & $16.2$ 
   & $0.083 \pm 0.067$ (8.3\%) & $0.23 \pm 0.006$ \\
 \hline
  1.0 & $13.7$ 
  & $0.092 \pm 0.065$ (9.2\%) & $0.24 \pm 0.009$ \\
 \hline
  2.0 & $12.0$ 
  & $0.121 \pm 0.064$ (12.1\%)  & $0.24 \pm 0.007$ \\
 \hline
  3.0 & $11.0$ 
  & $0.141 \pm 0.086$ (14.1\%) & $0.24 \pm 0.014$ \\
 \hline
 4.0 & $10.2$ 
 & $0.145 \pm 0.091$ (14.5\%) & $0.25 \pm 0.004$ \\
 \hline
  10.0 & $7.7$ 
  & $0.198 \pm 0.164$ (19.8\%) & $0.25 \pm 0.014$ \\
   \hline
  20.0 & $5.5$ 
  & $0.275 \pm  0.113$ (27.5\%) & $0.23 \pm 0.005$ \\
 \hline 
  \multicolumn{4}{| c |}{With joint torque, velocity and acceleration penalty}\\
  \hline
  - & $15.5$ 
  & $0.122 \pm 0.054 $ (12.2\%) &$0.27 \pm 0.008$ \\
 \hline
 
 \end{tabular}
 \label{table:energy}
  \vspace{2mm}
 \caption{Average Power vs. velocity error as a function of the power weight $c_E$.}
\end{table}

\subsection*{Studying the effect of the curriculum}

 \begin{figure}[!t]
     \centering
     \includegraphics[width=\linewidth]{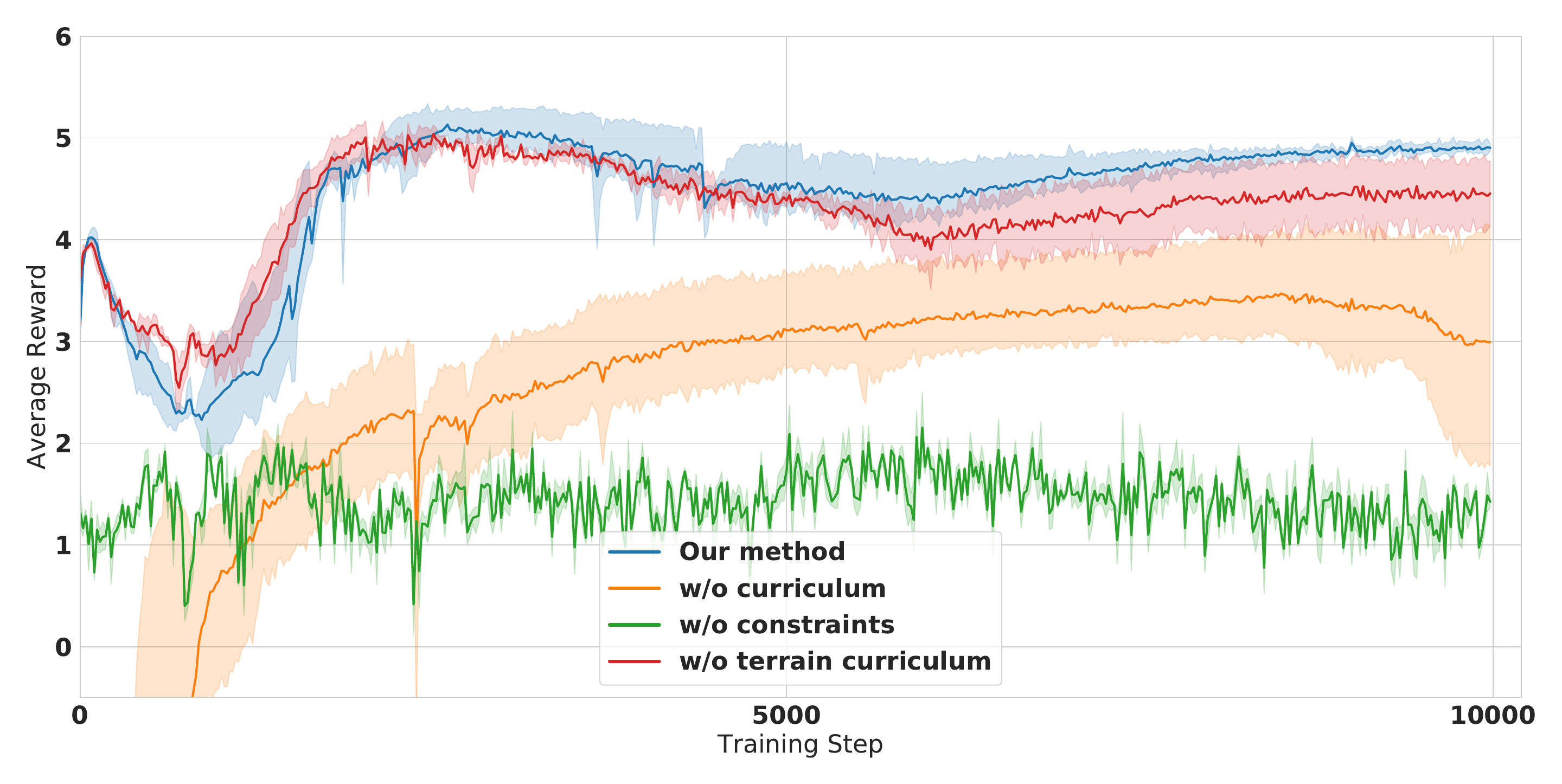}
     \caption{Plot of the average rewards over training steps for different setups. Each curve is averaged over three random seeds of the same experiment} 
     \label{fig:training}
 \end{figure}
 
\noindent In order to validate some of the choices made on the reward terms, curriculum and terrain curriculum, we run a set of ablation experiments.  Figure \ref{fig:training} shows the training curves that plot the average reward over the training steps for different setups. The blue curve shows our proposed method with the curriculum on the reward and terrain. The orange curve shows the experiments without using a curriculum. The red curve experiments the same reward curriculum but introduces the non-flat terrain from the start of training rather than at the end, as we propose. All the curves are averaged over three different runs of their respective experiments. 

 \begin{figure*}[!t]
     \centering
     \includegraphics[width=0.95\linewidth]{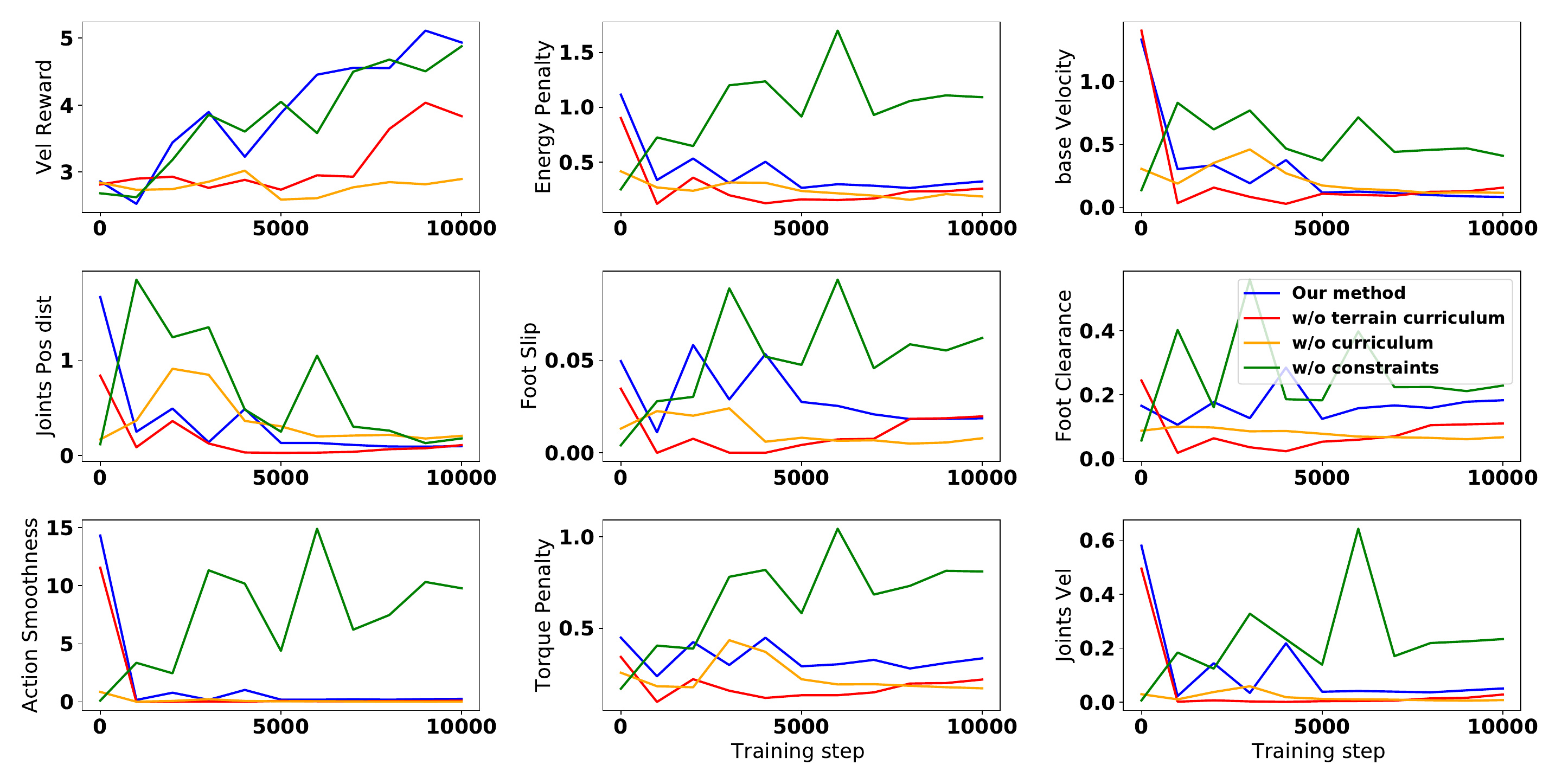}
     \caption{Plot of the separate reward penalties of the different curriculum settings over training steps} 
     \label{fig:penalties}
 \end{figure*}
 
Figure \ref{fig:training} shows that the proposed method with the curriculum outperforms the rest of the experiments in terms of the final average reward achieved and that the variance in the performance between the learned policies is low. This indicates that the learning is consistently reaching similar behaviours at the end of training. We also see that the experiments that use a curriculum achieve a higher reward at the start of training, which allows it to learn faster and reach a higher performance in the end. On the contrary, not using a curriculum results in slower learning, higher variance between runs and an asymptotically lower performance at the end of training.

The green curve is an experiment where an RL policy is trained only with the velocity tracking reward without the rest of the penalties. The curve plots the value of the reward with the penalty terms to show whether velocity tracking alone optimizes the other terms. As we see, the average reward performance for that experiment is very low, even though we observe that the velocity tracking term for these experiments is fully optimized.

Figure \ref{fig:penalties} shows the values of the individual reward terms for the same ablation experiments during the training process weighted by their chosen coefficients. The plot displays the average rewards achieved over three random seeds for each experiment.  The objective is to maximize the velocity tracking reward while minimizing the rest of the penalties. Our proposed training setup results in the best velocity tracking reward while optimizing the rest of the penalties. The experiments that do not use a reward curriculum (orange) or a terrain curriculum (red) optimize penalties but do not get a good performance over the main velocity tracking reward. This is in line with our motivation for designing the curriculum to learn the best trade-off between following the velocity and respecting the penalties.  We notice that the experiment trained on only tracking the reward (green) is able to maximize the velocity tracking term, however it doesn't respect any penalty terms. 

\begin{figure}[!t]
     \centering
     \includegraphics[width=0.6\linewidth]{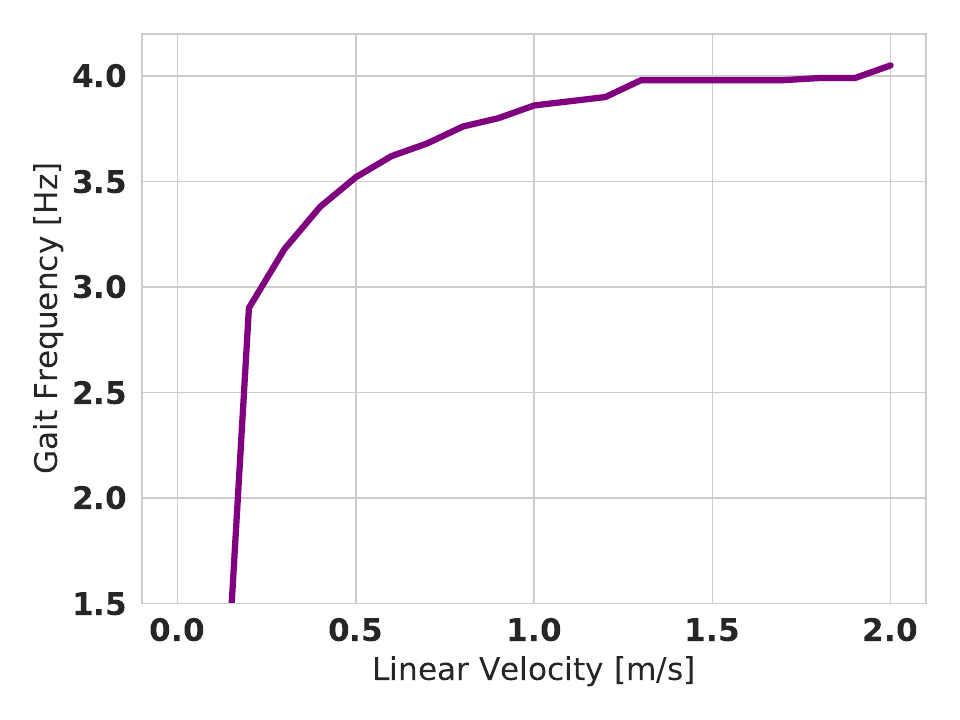}
     \caption{Plot of the gait frequency as a function of the linear velocity command}
     \label{fig:freq}
 \end{figure}
 
\subsection*{Gait frequency}
\noindent One of the desired features to have in a controller is the ability to adapt the gait frequency based on the velocity command. We show in our work that using RL, we can learn controllers that adapt their frequencies online. Using Fast Fourier Transform analysis (FFT) on the trajectory of the joint angles of the robot, we are able to deduce the frequency of the gait. Figure~\ref{fig:freq} shows the value of the frequency as a function of the linear velocity command. We see a proportional relationship between the velocity and gait frequency.  This behaviour emerges naturally during learning and is not hand-designed. This is an interesting result because adapting the gait frequency to velocity is not that straightforward to obtain through MPC-based controllers.

\subsection*{Comment on the policy transfer to Solo12}  

\noindent As explained earlier, random uniform noise was added to the robot dynamics and state observations during training. This noise was progressively inserted through the curriculum factor $k_{c,noise}$, starting with noiseless simulations and increasing the noise magnitude as the training progressed. The goal was to prepare the policy network for sim-to-real transfer so that, once deployed on a real Solo12, it would still produce robust behavior even if the model did not perfectly fit the system. Such discrepancy is inevitable since different motors have slightly different characteristics that vary as coils get warmer, and the model does not include joint friction, its inertia matrices are not perfectly accurate, etc.

Despite these inevitable model inaccuracies, the  policies were successfully transferred on the very first try. Even though Solo12 is a lightweight small robot, we were able to run it with our learned policies on various terrains, i.e., indoors, outdoors on grass and pebbles and on an ascending and descending slopes that are relatively steep considering the size of the robot\footnote{Video of the deployed policies on the real robot: https://youtu.be/t-67qBxNyZI}. These results show the robustness of the proposed control approach with respect to model variations. The transfer did not require learning an actuator model, as done in other works\cite{Hwangbo2019}, or modeling the actuation dynamics to include a bandwidth limitation through a low pass filter on the torques. This demonstrates how a simple randomization during training is enough for direct transfer to Solo12, probably by virtue of the fast dynamics of this robot (lightweight quadruped powered by low inertia actuators with high bandwidth) which leads to a limited sim-to-real gap. 
This all makes the Solo platform an attractive choice for deploying RL schemes.


\section*{Conclusion}
\label{sec:conclusion}
\noindent We presented an end-to-end approach for learning controllers for the Solo12 quadruped robot. We described the training method in detail with the choice of state space, action space and reward function along with the curriculum strategy and domain/dynamic randomization method for learning transferable policies for following 3D velocity commands. We presented results for the velocity tracking and energy loss. 
Numerous experimental tests on the real robot have shown that robust locomotion policies with different energy profiles can be learned by randomizing the weights of the power loss variables.
Based on this work and previous publications, we plan to conduct a large scale study in the near future to compare the potential of current model-based and  RL-based controllers on Solo12. 


\bibliography{sample}

@article{Miki2022,
author = {Takahiro Miki  and Joonho Lee  and Jemin Hwangbo  and Lorenz Wellhausen  and Vladlen Koltun  and Marco Hutter },
title = {Learning robust perceptive locomotion for quadrupedal robots in the wild},
journal = {Science Robotics},
volume = {7},
number = {62},
pages = {eabk2822},
year = {2022},

}

@article{Hwangbo2019,
author = {Hwangbo, Jemin and Lee, Joonho and Dosovitskiy, Alexey and Bellicoso, Dario and Tsounis, Vassilios and Koltun, Vladlen and Hutter, Marco},
journal = {Science Robotics},
number = {26},
title = {{Learning agile and dynamic motor skills for legged robots}},
volume = {4},
year = {2019}
}

@article{Lee2020,
author = {Lee, Joonho and Hwangbo, Jemin and Wellhausen, Lorenz and Koltun, Vladlen and Hutter, Marco},
journal = {Science Robotics},
number = {47},
title = {{Learning quadrupedal locomotion over challenging terrain}},
volume = {5},
year = {2020}
}

@inproceedings{Kumar2021,
  author    = {Ashish Kumar and
               Zipeng Fu and
               Deepak Pathak and
               Jitendra Malik},
  editor    = {Dylan A. Shell and
               Marc Toussaint and
               M. Ani Hsieh},
  title     = {{RMA:} Rapid Motor Adaptation for Legged Robots},
  booktitle = {Robotics: Science and Systems XVII, Virtual Event, July 12-16, 2021},
  year      = {2021},
  url       = {https://doi.org/10.15607/RSS.2021.XVII.011},
  doi       = {10.15607/RSS.2021.XVII.011},
  bibsource = {dblp computer science bibliography, https://dblp.org}
}

@article{Ji2022,
archivePrefix = {arXiv},
arxivId = {2202.05481v2},
author = {Ji, Gwanghyeon and Mun, Juhyeok and Kim, Hyeongjun and Hwangbo, Jemin},
journal = {IEEE Robotics and Automation Letters},
month = feb,
number = {2},
pages = {4630--4637},
publisher = {Institute of Electrical and Electronics Engineers Inc.},
title = {{Concurrent Training of a Control Policy and a State Estimator for Dynamic and Robust Legged Locomotion}},
volume = {7},
year = {2022}
}

@article{Li2021,
archivePrefix = {arXiv},
arxivId = {2103.14295},
author = {Li, Zhongyu and Cheng, Xuxin and Peng, Xue Bin and Abbeel, Pieter and Levine, Sergey and Berseth, Glen and Sreenath, Koushil},
isbn = {9781728190778},
issn = {10504729},
journal = {Proceedings - IEEE International Conference on Robotics and Automation},
pages = {2811--2817},
publisher = {Institute of Electrical and Electronics Engineers Inc.},
title = {{Reinforcement Learning for Robust Parameterized Locomotion Control of Bipedal Robots}},
volume = {2021-May},
year = {2021}
}

@article{Hoyt1981,
author = {Hoyt, Donald F. and Taylor, C. Richard},
issn = {00280836},
journal = {Nature},
month = jul,
number = {5820},
pages = {239--240},
publisher = {Nature Publishing Group},
title = {{Gait and the energetics of locomotion in horses}},
volume = {292},
year = {1981}
}

@book{Suttonbook,
  author = {Sutton, Richard S. and Barto, Andrew G.},
  edition = {Second},
  publisher = {The MIT Press},
  title = {Reinforcement Learning: An Introduction},
  year = {2018 }
}

@article{raisim,
  title={Per-contact iteration method for solving contact dynamics},
  author={Hwangbo, Jemin and Lee, Joonho and Hutter, Marco},
  journal={IEEE Robotics and Automation Letters},
  url="www.raisim.com",
  volume={3},
  number={2},
  pages={895--902},
  year={2018},
  publisher={IEEE}
}

@INPROCEEDINGS{Hutter2016,
  author={Hutter, Marco and Gehring, Christian and Jud, Dominic and Lauber, Andreas and Bellicoso, C. Dario and Tsounis, Vassilios and Hwangbo, Jemin and Bodie, Karen and Fankhauser, Peter and Bloesch, Michael and Diethelm, Remo and Bachmann, Samuel and Melzer, Amir and Hoepflinger, Mark},
  booktitle={2016 IEEE/RSJ International Conference on Intelligent Robots and Systems (IROS)}, 
  title={ANYmal - a highly mobile and dynamic quadrupedal robot}, 
  year={2016},
  volume={},
  number={},
  pages={38-44},
  }

@inproceedings{Katz2019,
author = {Katz, Benjamin and Carlo, Jared DI and Kim, Sangbae},
booktitle = {Proceedings - IEEE International Conference on Robotics and Automation},
month = may,
pages = {6295--6301},
publisher = {Institute of Electrical and Electronics Engineers Inc.},
title = {{Mini cheetah: A platform for pushing the limits of dynamic quadruped control}},
volume = {2019-May},
year = {2019}
}

@article{Grimminger2019,
archivePrefix = {arXiv},
arxivId = {1910.00093v2},
author = {Grimminger, Felix and Meduri, Avadesh and Khadiv, Majid and Viereck, Julian and W{\"{u}}thrich, Manuel and Naveau, Maximilien and Berenz, Vincent and Heim, Steve and Widmaier, Felix and Flayols, Thomas and Fiene, Jonathan and Badri-Spr{\"{o}}witz, Alexander and Righetti, Ludovic},
journal = {IEEE Robotics and Automation Letters},
month = sep,
number = {2},
pages = {3650--3657},
publisher = {Institute of Electrical and Electronics Engineers Inc.},
title = {{An Open Torque-Controlled Modular Robot Architecture for Legged Locomotion Research}},
volume = {5},
year = {2019}
}

@article{Semini2011,
author = {Claudio Semini and
          Nikos G Tsagarakis and 
          Emanuele Guglielmino and 
          Michele Focchi and 
          Ferdinando Cannella and 
          D G Caldwell},
title ={Design of HyQ – a hydraulically and electrically actuated quadruped robot},
journal = {Proceedings of the Institution of Mechanical Engineers, Part I: Journal of Systems and Control Engineering},
volume = {225},
number = {6},
pages = {831-849},
year = {2011},
}

@MISC{unitree,
    title =    {{Unitree Robotics}},
    author = {Xingxing Wang},
    howpublished = {\url{https://www.unitree.com/}}
}

@MISC{bostondyn,
    title = {{Spot Mini}},
    author = {Boston Dynamics},
    howpublished = {\url{https://www.bostondynamics.com/spot/}}
}

@misc{Kim2019,
author = {Kim, Donghyun and Carlo, Jared Di and Katz, Benjamin and Bledt, Gerardo and Kim, Sangbae},
title = {{Highly dynamic quadruped locomotion via whole-body impulse control and model predictive control}},
year = {2019}
}

@inproceedings{DiCarlo2018,
author = {{Di Carlo}, Jared and Wensing, Patrick M. and Katz, Benjamin and Bledt, Gerardo and Kim, Sangbae},
booktitle = {IEEE International Conference on Intelligent Robots and Systems},
title = {{Dynamic Locomotion in the MIT Cheetah 3 Through Convex Model-Predictive Control}},
year = {2018}
}

@inproceedings{Leziart2021,
  TITLE = {{Implementation of a Reactive Walking Controller for the New Open-Hardware Quadruped Solo-12}},
  AUTHOR = {L{\'e}ziart, Pierre-Alexandre and Flayols, Thomas and Grimminger, Felix and Mansard, Nicolas and Sou{\`e}res, Philippe},
  BOOKTITLE = {{2021 IEEE International Conference on Robotics and Automation - ICRA}},
  ADDRESS = {Xi'an, China},
  HAL_LOCAL_REFERENCE = {Rapport LAAS n{\textdegree} 20373},
  YEAR = {2021},
  MONTH = May,
  HAL_ID = {hal-03052451},
  HAL_VERSION = {v1},
}

@article{Bellicoso2018,
author = {Bellicoso, C. Dario and Jenelten, Fabian and Gehring, Christian and Hutter, Marco},
journal = {IEEE Robotics and Automation Letters},
month = jul,
number = {3},
pages = {2261--2268},
publisher = {Institute of Electrical and Electronics Engineers Inc.},
title = {{Dynamic Locomotion Through Online Nonlinear Motion Optimization for Quadrupedal Robots}},
volume = {3},
year = {2018}
}

@article{Winkler2015,
archivePrefix = {arXiv},
arxivId = {1904.03695},
author = {Winkler, Alexander W. and Mastalli, Carlos and Havoutis, Ioannis and Focchi, Michele and Caldwell, Darwin G. and Semini, Claudio},
isbn = {9781479969234},
issn = {10504729},
journal = {Proceedings - IEEE International Conference on Robotics and Automation},
month = jun,
number = {June},
pages = {5148--5154},
publisher = {Institute of Electrical and Electronics Engineers Inc.},
title = {{Planning and execution of dynamic whole-body locomotion for a hydraulic quadruped on challenging terrain}},
volume = {2015-June},
year = {2015}
}

@article{Levine2015,
archivePrefix = {arXiv},
arxivId = {1504.00702},
author = {Levine, Sergey and Finn, Chelsea and Darrell, Trevor and Abbeel, Pieter},
issn = {15337928},
journal = {Journal of Machine Learning Research},
month = apr,
pages = {1--40},
publisher = {Microtome Publishing},
title = {{End-to-End Training of Deep Visuomotor Policies}},
volume = {17},
year = {2015}
}

@misc{Kalashnikov2018,
archivePrefix = {arXiv},
arxivId = {1806.10293},
author = {Kalashnikov, Dmitry and Irpan, Alex and Pastor, Peter and Ibarz, Julian and Herzog, Alexander and Jang, Eric and Quillen, Deirdre and Holly, Ethan and Kalakrishnan, Mrinal and Vanhoucke, Vincent and Levine, Sergey},
month = jun,
title = {{QT-Opt: Scalable Deep Reinforcement Learning for Vision-Based Robotic Manipulation}},
year = {2018}
}

@article{Peters2008,
author = {Peters, Jan and Schaal, Stefan},
journal = {Neural Networks},
month = may,
number = {4},
pages = {682--697},
pmid = {18482830},
publisher = {Pergamon},
title = {{Reinforcement learning of motor skills with policy gradients}},
volume = {21},
year = {2008}
}

@article{Gullapalli1995,
author = {Gullapalli, Vijaykumar},
journal = {Robotics and Autonomous Systems},
month = oct,
number = {4},
pages = {237--246},
publisher = {North-Holland},
title = {{Skillful control under uncertainty via direct reinforcement learning}},
volume = {15},
year = {1995}
}

@article{Kober2010,
author = {Kober, Jens and M{\"{u}}lling, Katharina and Kr{\"{o}}mer, Oliver and Lampert, Christoph H. and Sch{\"{o}}lkopf, Bernhard and Peters, Jan},
journal = {Proceedings - IEEE International Conference on Robotics and Automation},
pages = {853--858},
title = {{Movement templates for learning of hitting and batting}},
year = {2010}
}

@article{Kalakrishnan2012,
author = {Kalakrishnan, Mrinal and Righetti, Ludovic and Pastor, Peter and Schaal, Stefan},
journal = {Proceedings of the 29th International Conference on Machine Learning, ICML 2012},
month = sep,
pages = {4639--4644},
publisher = {IEEE},
title = {{Learning force control policies for compliant robotic manipulation}},
volume = {1},
year = {2012}
}

@article{Benbrahim1997,
author = {Benbrahim, Hamid and Franklin, Judy A.},
journal = {Robotics and Autonomous Systems},
month = dec,
number = {3-4},
pages = {283--302},
publisher = {North-Holland},
title = {{Biped dynamic walking using reinforcement learning}},
volume = {22},
year = {1997}
}

@article{Kohl2004,
author = {Kohl, Nate and Stone, Peter},
journal = {Proceedings - IEEE International Conference on Robotics and Automation},
number = {3},
pages = {2619--2624},
publisher = {Institute of Electrical and Electronics Engineers Inc.},
title = {{Policy gradient reinforcement learning for fast quadrupedal locomotion}},
volume = {2004},
year = {2004}
}

@article{Mnih2015,
author = {Mnih, Volodymyr and Kavukcuoglu, Koray and Silver, David and Rusu, Andrei A. and Veness, Joel and Bellemare, Marc G. and Graves, Alex and Riedmiller, Martin and Fidjeland, Andreas K. and Ostrovski, Georg and Petersen, Stig and Beattie, Charles and Sadik, Amir and Antonoglou, Ioannis and King, Helen and Kumaran, Dharshan and Wierstra, Daan and Legg, Shane and Hassabis, Demis},
journal = {Nature 2015 518:7540},
month = feb,
number = {7540},
pages = {529--533},
pmid = {25719670},
publisher = {Nature Publishing Group},
title = {{Human-level control through deep reinforcement learning}},
volume = {518},
year = {2015}
}

@misc{Koutnik2013,
address = {New York, New York, USA},
author = {Koutn{\'{i}}k, Jan and Cuccu, Giuseppe and Schmidhuber, J{\"{u}}rgen and Gomez, Faustino},
journal = {GECCO 2013 - Proceedings of the 2013 Genetic and Evolutionary Computation Conference},
pages = {1061--1068},
publisher = {ACM Press},
title = {{Evolving large-scale neural networks for vision-based reinforcement learning}},
year = {2013}
}

@misc{Mnih2013,
 archivePrefix = {arXiv},
arxivId = {1312.5602},
author = {Mnih, Volodymyr and Kavukcuoglu, Koray and Silver, David and Graves, Alex and Antonoglou, Ioannis and Wierstra, Daan and Riedmiller, Martin},
month = dec,
title = {{Playing Atari with Deep Reinforcement Learning}},
year = {2013}
}

@inproceedings{Levine2013,
author = {Levine, Sergey and Koltun, Vladlen},
booktitle = {30th International Conference on Machine Learning, ICML 2013},
number = {PART 2},
pages = {1038--1046},
publisher = {JMLR.org},
series = {ICML'13},
title = {{Guided policy search}},
year = {2013}
}

@misc{Schulman2017,
author = {Schulman, John and Wolski, Filip and Dhariwal, Prafulla and Radford, Alec and Klimov, Oleg},
booktitle = {arXiv},
issn = {23318422},
title = {{Proximal policy optimization algorithms}},
year = {2017}
}

@article{Schulman2015,
author = {Schulman, John and Levine, Sergey and Moritz, Philipp and Jordan, Michael and Abbeel, Pieter},
journal = {32nd International Conference on Machine Learning, ICML 2015},
month = feb,
pages = {1889--1897},
publisher = {International Machine Learning Society (IMLS)},
title = {{Trust region policy optimization}},
volume = {3},
year = {2015}
}

@article{Lecun2015,
author = {Lecun, Yann and Bengio, Yoshua and Hinton, Geoffrey},
journal = {Nature},
month = may,
number = {7553},
pages = {436--444},
pmid = {26017442},
publisher = {Nature Publishing Group},
title = {{Deep learning}},
volume = {521},
year = {2015}
}

@article{LeCun1998,
author = {LeCun, Yann and Bottou, L{\'{e}}on and Bengio, Yoshua and Haffner, Patrick},
journal = {Proceedings of the IEEE},
number = {11},
pages = {2278--2323},
title = {{Gradient-based learning applied to document recognition}},
volume = {86},
year = {1998}
}

@misc{Krizhevsky2012,
author = {Krizhevsky, Alex and Sutskever, Ilya and Hinton, Geoffrey E.},
booktitle = {Advances in Neural Information Processing Systems},
pages = {1097--1105},
title = {{ImageNet classification with deep convolutional neural networks}},
volume = {2},
year = {2012}
}

@article{Peng2016,
author = {Peng, Xue Bin and van de Panne, Michiel},
journal = {Proceedings - SCA 2017: ACM SIGGRAPH / Eurographics Symposium on Computer Animation},
month = nov,
publisher = {Association for Computing Machinery, Inc},
title = {{Learning Locomotion Skills Using DeepRL: Does the Choice of Action Space Matter?}},
year = {2016}
}

@inproceedings{Bengio2009,
address = {New York, New York, USA},
author = {Bengio, Yoshua and Louradour, J{\'{e}}r̂ome and Collobert, Ronan and Weston, Jason},
booktitle = {ACM International Conference Proceeding Series},
pages = {1--8},
publisher = {ACM Press},
title = {{Curriculum learning}},
volume = {382},
year = {2009}
}

@article{FRENCH1999128,
title = {Catastrophic forgetting in connectionist networks},
journal = {Trends in Cognitive Sciences},
volume = {3},
number = {4},
pages = {128-135},
year = {1999},
author = {Robert M. French},
}

@article{stable-baselines3,
  author  = {Antonin Raffin and Ashley Hill and Adam Gleave and Anssi Kanervisto and Maximilian Ernestus and Noah Dormann},
  title   = {Stable-Baselines3: Reliable Reinforcement Learning Implementations},
  journal = {Journal of Machine Learning Research},
  year    = {2021},
  volume  = {22},
  number  = {268},
  pages   = {1-8},
}

@inproceedings{Fadini2021,
  TITLE = {{Computational design of energy-efficient legged robots: Optimizing for size and actuators}},
  AUTHOR = {Fadini, Gabriele and Flayols, Thomas and del Prete, Andrea and Mansard, Nicolas and Sou{\`e}res, Philippe},
  BOOKTITLE = {{IEEE International Conference on Robotics and Automation (ICRA 2021)}},
  YEAR = {2021}}

@inproceedings{Kingma2014,
  author    = {Diederik P. Kingma and
               Jimmy Ba},
  editor    = {Yoshua Bengio and
               Yann LeCun},
  title     = {Adam: {A} Method for Stochastic Optimization},
  booktitle = {3rd International Conference on Learning Representations, {ICLR} 2015,
               San Diego, CA, USA, May 7-9, 2015, Conference Track Proceedings},
  year      = {2015},
}

@inproceedings{li2023cassi,
  title = {Versatile Skill Control via Self-supervised Adversarial Imitation of Unlabeled Mixed Motions},
  author = {Li, C. and Blaes, S. and Kolev, P. and Vlastelica, M. and Frey, J. and Martius, G.},
  booktitle = {Proceedings of the IEEE International Conference on Robotics and Automation (ICRA)},
  month = may,
  year = {2023},
  doi = {},
  month_numeric = {5}
}

@misc{Fuchioka2022-opt-mimic,
  doi = {10.48550/ARXIV.2210.01247},
  author = {Fuchioka, Yuni and Xie, Zhaoming and van de Panne, Michiel},
  title = {OPT-Mimic: Imitation of Optimized Trajectories for Dynamic Quadruped Behaviors},
  year = {2022},
  copyright = {Creative Commons Attribution 4.0 International}
}

\section*{Author contributions statement}

M.A. developed the training method and code. M.A, P.S., T.S wrote the main manuscript text, T.F, P.A.L. prepared the real robot experiments and coded the interface with the robot platform, T.F. edited the video submission, J.P. reviewed the method and experimental results. All authors reviewed the manuscript.






\end{document}